\documentclass[11pt,a4paper]{article}
\usepackage{times}
\usepackage{latexsym}
\usepackage[utf8]{inputenc}
\usepackage{lipsum}
\usepackage{enumitem}
\usepackage{subfigure}

\usepackage{amsmath}
\usepackage{cite}

\usepackage{url}

\usepackage{graphicx}
\usepackage[hyperref]{naaclhlt2019}

\aclfinalcopy 
\title{Neural Vector Conceptualization for Word Vector Space Interpretation}

\author{\textbf{Robert Schwarzenberg*, Lisa Raithel*, David Harbecke} \\ German Research Center for Artificial Intelligence (DFKI), Berlin, Germany\\
{\tt \{firstname.lastname\}@dfki.de}
}

\date{}

\newcommand\blfootnote[1]{%
  \begingroup
  \renewcommand\thefootnote{}\footnote{#1}%
  \addtocounter{footnote}{-1}%
  \endgroup
}

\begin{document}
\maketitle

\begin{abstract}
Distributed word vector spaces are considered hard to interpret which hinders the understanding of natural language processing (NLP) models. In this work, we introduce a new method to interpret arbitrary samples from a word vector space. To this end, we train a neural model to conceptualize word vectors, which means that it activates higher order concepts it recognizes in a given vector.  Contrary to prior approaches, our model operates in the original vector space and is capable of learning non-linear relations between word vectors and concepts. Furthermore, we show that it produces considerably less entropic concept activation profiles than the popular cosine similarity.\blfootnote{* Shared first authorship.}
\end{abstract}

\section{Introduction}

In the vast majority of state-of-the-art NLP models, as for instance in translation models \citep{bojar_findings_2018} or text classifiers \citep{howard_universal_2018}, language is represented in distributed vector spaces. 
Using distributed representations comes at the price of low interpretability as they are generally considered uninterpretable, without further means \citep{levy_dependency-based_2014, montavon_methods_2018}. 
In this work, we address this lack of interpretability with \textit{neural vector conceptualization} (NVC), a neural mapping from a word vector space to a concept space (e.g.~``chair'' should activate the concept ``furniture'').

Using concepts to interpret distributed vector representations of language is inspired by the finding that ``humans understand languages through multi-step cognitive processes which involves building rich models of the world and making multi-level generalizations from the input text'' \citep{shalaby2019learning}. We are not the first, however, to utilize concepts for this purpose.

\citet{koc_imparting_2018}, for instance, modify the objective function of \texttt{GloVe} \citep{pennington_glove_2014} to align semantic concepts with word vector dimensions to create an interpretable space. Their method does not, however, offer an interpretation of vectors in the original space. 

\citet{senel_semantic_2018}, in contrast, do offer an interpretation of the original space:
They propose a mapping of word vector dimensions to concepts. This mapping, however, is linear and consequently, their method is incapable of modeling non-linear relations.

Our method offers an interpretation of the original space and is capable of modeling non-linear relations between the word and the concept space.
Furthermore, arguably, we interpret vectors similar to how a neural NLP model would, because a neural NLP model lies at the heart of our method.
In addition, by design, our model is able to conceptualize random continuous samples, drawn from the word vector space.

This is particularly important as word vectors are sparse in their vector space and vectors without a word representative do not have intrinsic meaning. This hinders adapting methods from vision, such as activation maximization \citep{simonyan_deep_2013} or generative adversarial networks \citep{goodfellow_generative_2014}, as in NLP these methods potentially produce vectors without word representations. 

For introspection, one could map any vector onto its nearest neighbor with a word representative. However, nearest neighbor search does not necessarily find the closest semantic representative in the vector space \citep{schnabel_evaluation_2015}. Moreover, we show that concept activation profiles produced with nearest neighbor search tend to be considerably more entropic than the activation profiles our method returns. 

\section{Method}
\label{sec:Method}

For NVC, we propose to train a neural model to map word vectors onto associated concepts. More formally, the model should learn a meaningful mapping 
\begin{equation}
    f: {\rm I\!R}^{d} \rightarrow {\rm I\!R}^{|C|}
\end{equation}
where $d$ denotes the number of word vector dimensions and $C$ is a set of concepts. The training objective should be a multi-label classification to account for instances that belong to more than one concept (e.g.~``chair'' should also activate ``seat''). 

For the training, we need to make two basic choices:  
\begin{enumerate}[topsep=0pt,itemsep=-1ex,partopsep=1ex,parsep=1ex]
    \item We need a ground truth concept knowledge base that provides the concepts a training instance should activate and
    \item we need to choose a model architecture appropriate for the task.
\end{enumerate}
In the following, we motivate our choices.

\subsection{Ground Truth Concept Knowledgebase}

As a ground truth concept knowledge base we chose the Microsoft Concept Graph (MCG), which is built on top of Probase, for the following reasons:
\begin{enumerate}
    \item \citet{wu_probase_2012} convincingly argue that with Probase they built a universal taxonomy that is more comprehensive than other existing candidates, such as for example, \textit{Freebase} \citep{bollacker_freebase_2008}.
    \item Furthermore, Probase is huge. The core taxonomy contains about 5.38 million concepts, 12.5 million unique instances, and 85.1 million \textit{isA} relations. This allows our model to illuminate the word vector space from many angles.
    \item Instance-concept relations are probabilistic in the MCG: For (instance, concept) tuples a $rep$ score can be retrieved. The $rep$ score describes the ``representativeness'' of an instance for a concept, and vice versa. According to the MCG, for example, the instance ``chair'' is a few thousand times more representative for the concept ``furniture'' than is the instance ``car.'' During training, we exploit the $rep$ scores to retrieve representative target concepts for a training instance. 
\end{enumerate}

The scores are based on the notion of Basic Level Concepts (BLC) which were first introduced by \citet{rosch_basic_1976}, as part of Prototype Theory.
A basic level concept is a concept on which all people of the same culture consciously or unconsciously agree. For instance, according to Prototype Theory, most humans would categorize a ``wood frog'' simply as a ``frog.'' ``Wood frog'' is a representative instance of the concept ``frog.''

Aiming to provide an approach to the computation of the BLC of an instance $i$ in the MCG, \citet{wang_inference_2015} combine pointwise mutual information (PMI) 
with co-occurrence counts of concept $c$ and instance $i$. The authors compute the ``representativeness" of an instance $i$ for a concept $c$ as

\begin{equation}
rep(i, c) = P (c|i) \cdot P (i|c).
\end{equation}
By taking the logarithm of the $rep$ score, we can isolate the involvement of PMI:
\begin{equation}\label{eq:log_rep}
log~rep(i, c) - log~P(i, c) = PMI(i, c).
\end{equation}
In doing so, the authors boost concepts in the middle of the taxonomy (the basic level concepts) while reducing extreme values leading to super- or subordinate concepts. 
To find the BLC of a single instance, \citet{wang_inference_2015} maximize over the $rep$ value of all concepts associated with $i$.

To train our model, for a training instance $i$, we collect all concepts for which $rep(i,c)$\footnote{We computed the rep values ourselves as we only acquired a count-based version of the graph.} is above a certain threshold and use them as the target labels for $i$. We discard concepts that have very few instances above a threshold $rep$ value in the graph. 

\begin{figure*}[!ht]
\centering
\includegraphics[scale=.7]{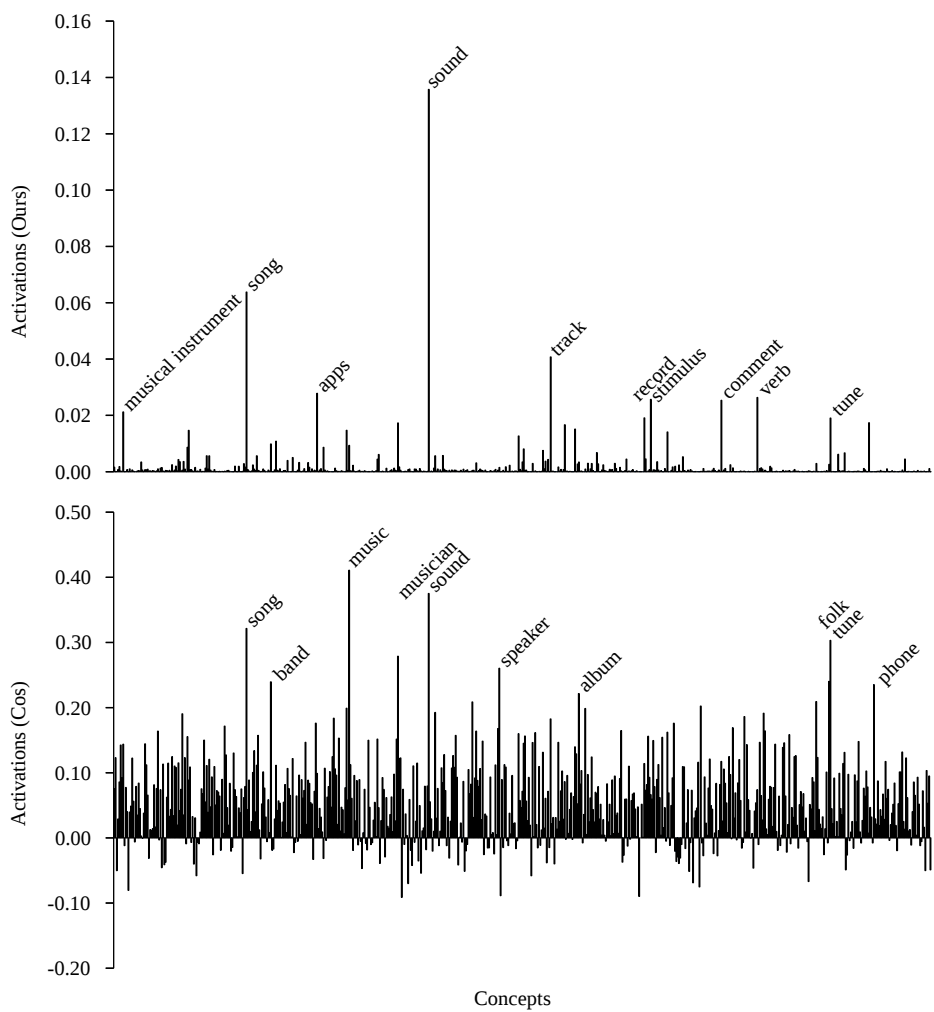}
\caption{Vector interpretations of the word vector of ``listening" with 637 concepts. Top: Neural vector conceptualization (our method, 10 highest activations labelled). Bottom: Cosine similarity (baseline, 10 highest activations labelled). Both activation profiles are unnormalized.}
\label{fig:listening}
\end{figure*}

\subsection{Model}
During training, the model repeatedly receives a word vector instance as input and a multi-hot vector retrieved from the MCG as the target concept vector. Thus, it must identify concepts encoded in the word vector.

We do not see any sequentiality or recurrence in this task which is why we discarded recurrent and Transformer candidate models.
Concerning convolutional networks, we disregard small receptive fields because dimensional adjacency is semantically irrelevant in word vectors.
However, any convolutional network with a receptive field over the whole input vector is equivalent to a fully-connected (FC) feed-forward network.
Thus, we ultimately trained an FC feed-forward network to conceptualize vectors. 

\section{Experiments}
For a proof of concept, we chose the \texttt{word\-2\-vec} embedding \citep{mikolov_efficient_2013} as the word vector space to interpret.
Recently, contextualized representations, like \texttt{EL\-Mo} \citep{peters2018deep} and \texttt{BERT} \citep{devlin_bert_2018-1}, received increased attention.
Nevertheless, well-established global representations, such as \texttt{word2vec} remain highly relevant:
\texttt{ELMo} still benefits from using global embeddings as additional input and \texttt{BERT} trains its own global token embedding space.

The \texttt{word2vec} model and the MCG are based on different corpora.
As a consequence of using data from two different sources, we sometimes needed to modify MCG instances to match the \texttt{word2vec} vocabulary.

We filtered the MCG for concepts that have at least 100 instances with a $rep$ value of at least $-$10. 
This leaves 637 concepts with an average of 184 instances per concept and gives a class imbalance of 524 negative samples for every positive sample.

With the obtained data, we trained a three-layer FC network to map word vectors onto their concepts in the MCG. The model returns independent sigmoid activations for each concept.
We trained with categorical cross entropy and applied weights regularization with a factor of $10^{-7}$.
For all experiments, we optimized parameters with the ADAM optimizer \citep{kingma_adam_2014}.\footnote{Our experiments are open source and can be replicated out of the box: \url{https://github.com/dfki-nlp/nvc}.}

\begin{figure*}
    \centering
    \includegraphics[width=\textwidth]{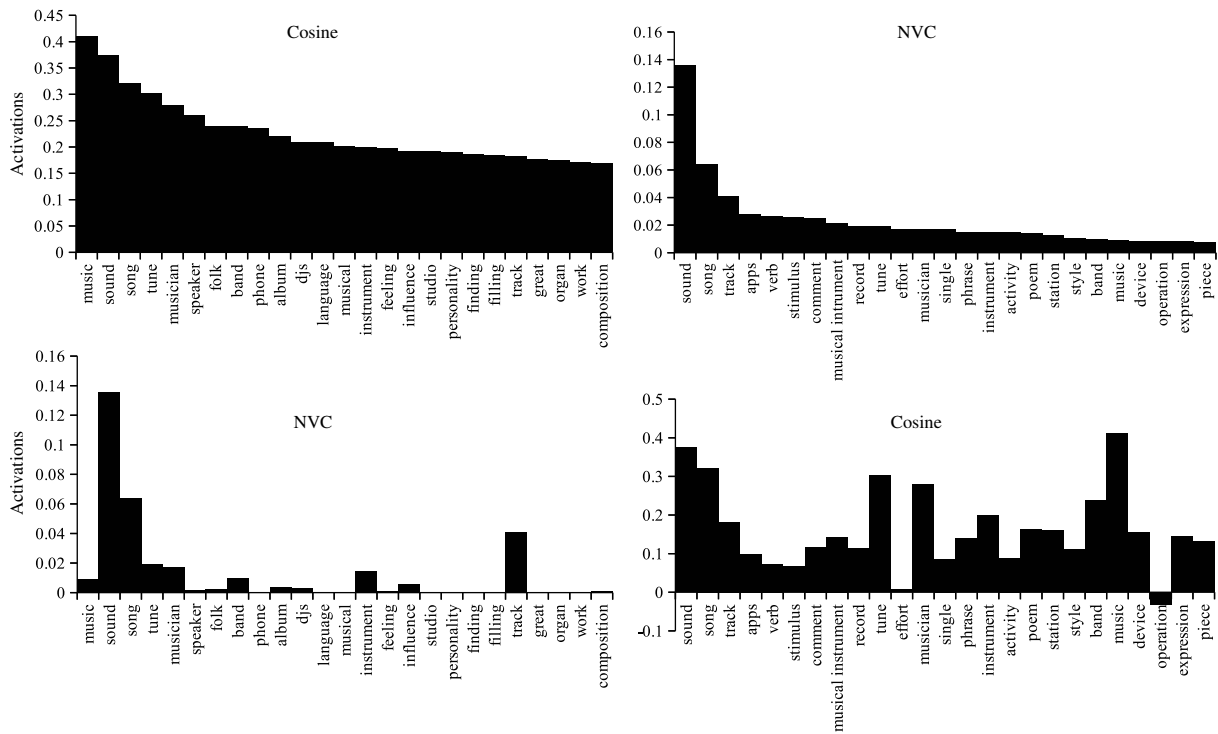}
    \caption{Concept activations for the instance ``listening.'' Upper left: Top 25 concepts according to cosine similarity. Bottom left: NVC activations of the same cosine top 25 concepts. Upper right: Top 25 concepts according to NVC. Bottom right: Cosine activations of the same NVC top 25 concepts.}
    \label{fig:top25}
\end{figure*}

To estimate task complexity, Table~\ref{tab:results} lists the precision, recall and F$_1$ scores that our model achieved on a fixed, randomly sampled test set that contained 10 \% of the data.
The table contains the weighted average scores accomplished for all concepts as well as the scores the model achieved for selected individual concepts, grouped semantically. 

Fig.~\ref{fig:listening} juxtaposes the NVC and the baseline activation profile of the word vector of ``listening'', which was not encountered during training. Several other NVCs can be found in the appendix (see Figs.~\ref{fig:mafioso}, \ref{fig:jealousy} and \ref{fig:berlin}) as well as selected concept activations of continuous samples (see Fig.~\ref{fig:listening_speaking_path}).

While Fig.~\ref{fig:listening} shows a global perspective of the activation profiles, Fig.~\ref{fig:top25} zooms in on the top 25 concepts, activated by the baseline method (first column) and our method (second column).

\begin{table}[!htb]
\centering
\begin{tabular}{c|cccc}
    \ & P & R & F & S \\
    \hline
    all concepts    &   0.43    &   0.16    &   0.22    &   9766 \\
    \hline
    province        &   0.81    &   0.81    &   0.81    &   36 \\
    district        &   0.79    &   0.62    &   0.69    &   78 \\
    island          &   0.96    &   0.38    &   0.54    &   64 \\
    locality        &	0.5     &	0.03    &	0.06    &	29 \\
    location        &   0       &   0       &   0       &   14 \\
    \hline
    choreographer   &	0.85    &   0.69    &	0.76    &	16 \\
    composer        &	0.8     &	0.66    &   0.72    &	61 \\
    artist          &   0.57    &   0.36    &   0.44    &   70 \\
    legend          &   0       &   0       &   0       &   33 \\
    \hline
    dish            &   0       &   0       &   0       &   34 \\
    meal            &   0       &   0       &   0       &   17 \\
    delicacy        &   0       &   0       &   0       &   11 \\
    salad           &   0       &   0       &   0       &   9  \\
\end{tabular}
\caption{Precision (P), recall (R), $F_1$ Score (F), and support (S) for all 637 concepts ($F_1$ Score weighted by support) and selected individual concepts. Class membership was determined by an activation threshold of 0.5.}
\label{tab:results}
\end{table}

\section{Discussion}
The weighted classification $F_1$ score is $0.22$ which suggests that the task is complex, probably due to the highly imbalanced data set. According to Table~\ref{tab:results}, however, $F_{1}$ scores vary significantly along individual concepts. While we observe a high score for \textit{province}, our model has difficulties classifying \textit{location}s, for instance. The same trend can be observed for \textit{choreographer}s and \textit{legend}s. What we see reflected in this table is the sharpness of concept boundaries. Arguably, the definition of a \textit{province} is sharper than that of \textit{location}. The same is true for \textit{choreographer} and \textit{legend}. We assume that the more precise a concept boundary, the higher the classification performance tends to be.  We cannot, however, offer an explanation for the poor classification performance on some other concepts, such as the last ones in Table~\ref{tab:results}. 

Fig.~\ref{fig:listening} (top) shows the NVC of ``listening'' with the top ten peaks labelled. For Table~\ref{tab:results}, a class membership was determined by an activation threshold of $0.5$ of the relevant output neuron.  Fig.~\ref{fig:listening} (top), however, illustrates that the model activates many meaningful concepts beneath this threshold and thus $0.5$ might not be appropriate to determine class membership.  

Some of the peaks are also reflected in the bottom plot of Fig.~\ref{fig:listening}, which depicts the activation profile of the cosine similarity baseline method. The most notable difference between our method and the baseline is that the latter produces much more entropic activation profiles. It is less selective than NVC as NVC deactivates many concepts. 

Fig.~\ref{fig:top25} (first column) shows that NVC indeed deactivates unrelated concepts, such as \textit{personality}, \textit{finding}, \textit{filling}, \textit{great}, and \textit{work} that, according to cosine similarity, are close to the instance ``listening.'' \textit{Speaker}, \textit{phone}, and \textit{organ} arguably are reasonable concepts and yet deactivated by NVC but NVC replaces them with more meaningful concepts, as can be seen in the upper right plot in Fig.~\ref{fig:top25}. Note that, contrary to NVC, the baseline method is not able to deactivate concepts that have close vectors in the word vector space, nor is it able to activate concepts that have vectors that are far from the input vector. 
Overall, a manual analysis suggests that the top 25 NVC concepts are more fitting than the top 25 cosine concepts.

\section{Related Work}
Concept knowledge bases such as the MCG exist because concepts are powerful abstractions of natural language instances that have been used for many downstream tasks, such as text classification \citep{song2011short}, ad-query similarity and query similarity \citep{kim2013context}, document similarity \citep{song-roth:2015:NAACL-HLT}, and semantic relatedness \citep{bekkali2019effective}. The approaches mentioned above all implement some form of text conceptualization (TC).

TC models the probability $P(c|I)$  of a concept $c$ being reflected in a set of observed natural language instances $I$ \citep{song2011short, shalaby2019learning}. This is also the objective function of the model we train and our interpretability method can thus be understood as an implementation of TC. 

Furthermore, besides the methods already discussed in the introduction, there is more research into the interpretability of language representations. \citet{adi2017}, for instance, also use auxiliary prediction tasks to analyse vector representations. However, they work on sentence level, not word level. Moreover, instead of retrieving concepts, they probe sentence length, word content conservation and word order conservation in the representation. 

An approach similar to ours was introduced by \citet{sommerauer2018firearms}. The authors investigate the kind of semantic information encoded in word vectors. To this end, they train a classifier that recognizes whether word vectors carry specific semantic properties, some of which can be regarded as concepts. 

\section{Conclusion \& Future Work}
\label{ssec:further_work}
We introduced neural vector conceptualization as a means of interpreting continuous samples from a word vector space. We demonstrated that our method produces considerably less entropic concept activation profiles than the cosine similarity measure. For an input word vector, NVC activated meaningful concepts and deactivated unrelated ones, even if they were close in the word vector space. 

Contrary to prior methods, by design, NVC operates in the original language space and is capable of modeling non-linear relations  between language instances and concepts. Furthermore, our method is flexible: At the heart of it lies a neural NLP model that we trained on an instance-concept ground truth that could be replaced by another one.

In the future, we would like to extend NVC to contextualized representations. We consider this non-trivial because it may not be possible to directly apply the current instance-concept ground truth to contextualized instances, in particular if they are represented by sub-word embeddings. 

\section*{Acknowledgements}
This research was partially supported by the German Federal Ministry of Education and Research through the project DEEPLEE (01IW17001). We would also like to thank the anonymous reviewers for their feedback and Leonhard Hennig for data and feedback.

\bibliography{revised}
\bibliographystyle{acl_natbib}
\appendix

\section{NVCs}
\label{sec:appendix}

\begin{figure}[!htb]
\centering
\includegraphics[width=0.5\textwidth]{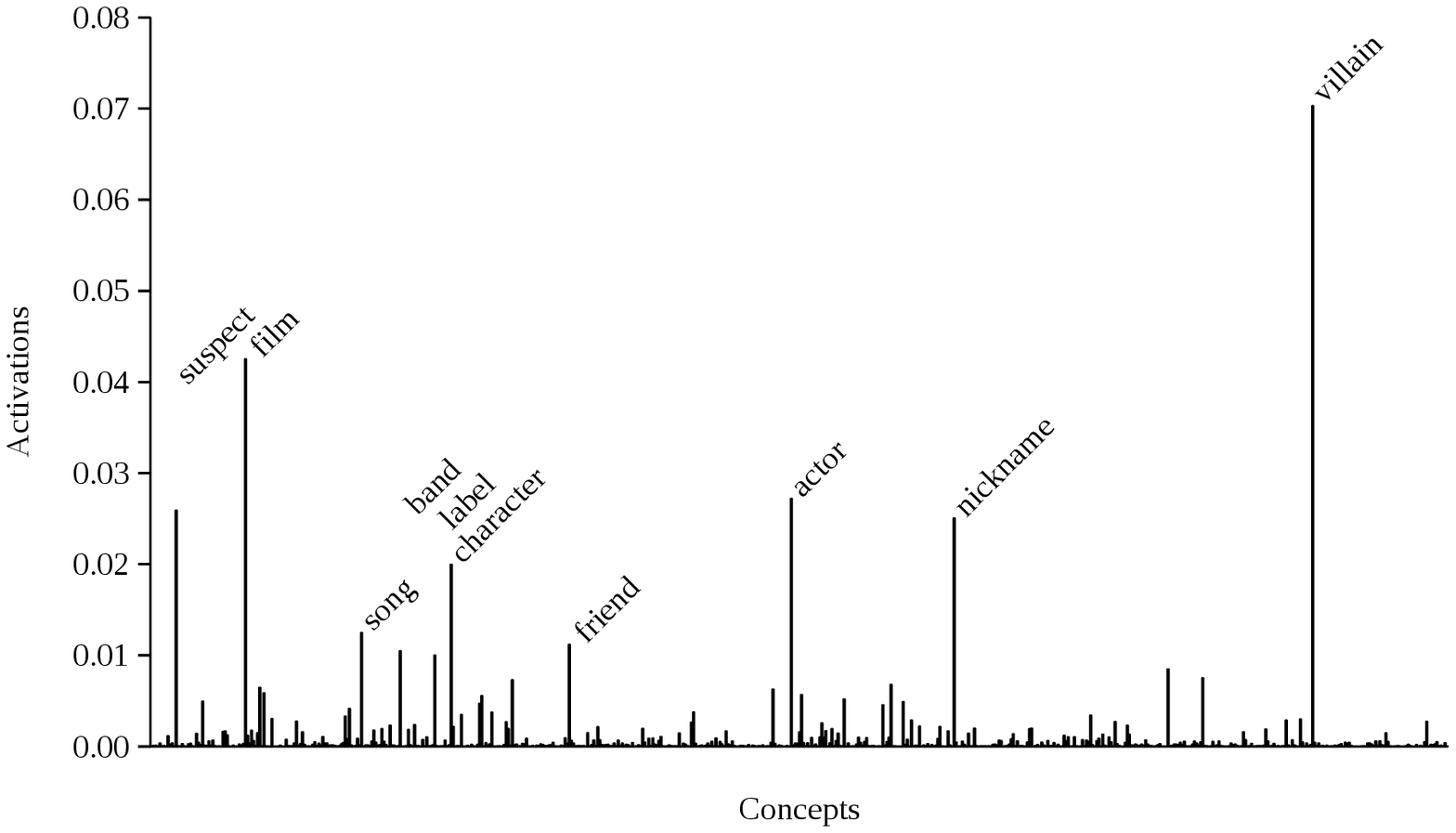}

\caption{NVC of the word vector for ``mafioso'' (the instance was not encountered during training).}
\label{fig:mafioso}
\end{figure}

\begin{figure}[!htb]
\centering
\includegraphics[width=0.5\textwidth]{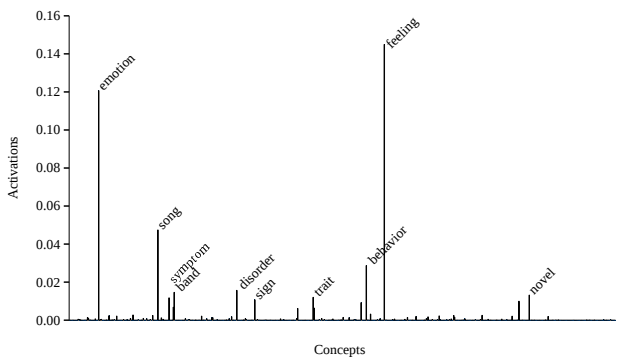}

\caption{NVC of the word vector for ``Jealousy'' (the instance was not encountered during training).}
\label{fig:jealousy}
\end{figure}

\begin{figure}[!h]
\centering
\includegraphics[width=0.5\textwidth]{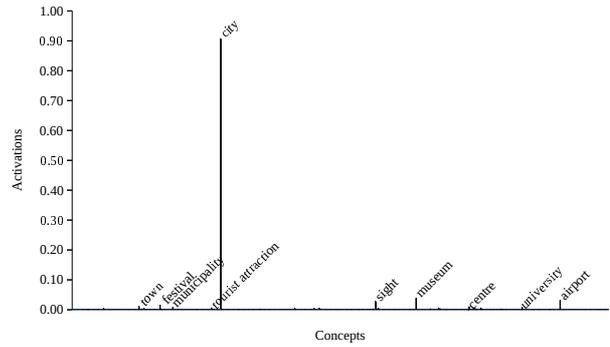}

\caption{NVC of the word vector for ``Berlin'' (the instance was not encountered during training).}
\label{fig:berlin}
\end{figure}

\section{Concept Activations for Continuous Samples}
\begin{figure}[!h]
\centering
\includegraphics[width=0.5\textwidth]{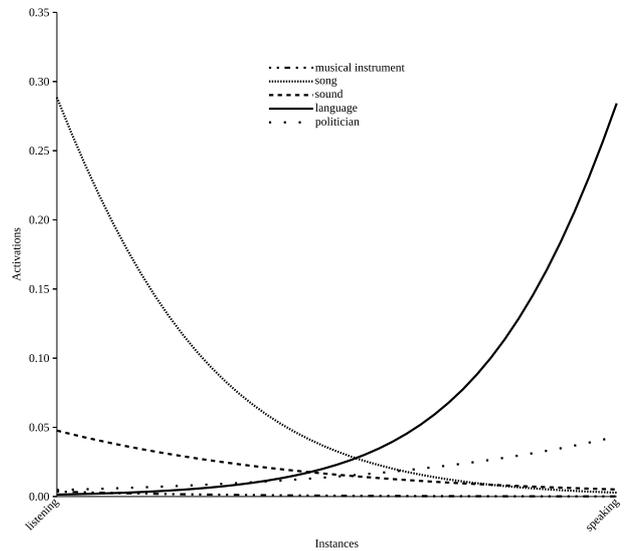}
\caption{Concept activations of five selected concepts of word vectors sampled on the path between the instances ``listening'' and ``speaking''. Note the steady, non-oscillating paths between the instances.}
\label{fig:listening_speaking_path}
\end{figure}

\end{document}